\documentclass[sigconf]{acmart}

\usepackage{booktabs} 

\setcopyright{rightsretained}

\acmDOI{10.475/123_4}

\acmISBN{123-4567-24-567/08/06}

\acmConference[KDD'17]{ACM SIGKDD conference}{2017}{Halifax, Nova Scotia Canada} 
\acmYear{2017}
\copyrightyear{2017}

\acmPrice{15.00}

\usepackage{color}
\usepackage{bm}
\usepackage{hyperref}
\usepackage{amsmath,amsfonts,amsthm}
\usepackage{blindtext}
\usepackage{listings}

\newif\ifdebug
\debugtrue

\ifdebug
\newcommand{\chris}[1]{{\color{red}{\bf\sf [CJF: #1]}}}
\newcommand{\junz}[1]{{\color{blue}{\bf\sf [ZJ: #1]}}}
\newcommand{\lkw}[1]{{\color{cyan}{\bf\sf [LKW: #1]}}}
\newcommand{\cwg}[1]{{\color{green}{\bf\sf [CWG: #1]}}}
\else
\newcommand{\chris}[1]{{\color{red}{}}}
\newcommand{\junz}[1]{{\color{blue}{}}}
\newcommand{\lkw}[1]{{\color{cyan}{}}}
\newcommand{\cwg}[1]{{\color{green}{}}}
\fi

\newcommand{\vect}[1]{\boldsymbol{\mathbf{#1}}}

\newcommand{\Dir}{\mbox{Dir}}
\newcommand{\Mult}{\mbox{Mult}}

\newcommand{\alphav}{\vect\alpha}
\newcommand{\betav}{\vect\beta}
\newcommand{\gammav}{\vect\gamma}

\newcommand{\thetav}{\vect\theta}

\newcommand{\piv}{\vect\pi}

\newcommand{\phiv}{\vect\phi}

\newcommand{\cv}{\vect c}

\newcommand{\ev}{\vect e}
\newcommand{\fv}{\vect f}

\newcommand{\mv}{\vect m}

\newcommand{\wv}{\vect w}

\newcommand{\zv}{\vect z}

\newcommand{\Cv}{\vect C}

\newcommand{\Wv}{\vect W}

\usepackage{subfigure}

\begin{document}
\title{Scalable Inference for Nested Chinese Restaurant Process Topic Models}

\author{Jianfei Chen$^{\dag}$, Jun Zhu$^{\dag}$, Jie Lu$^\ddag$, and Shixia Liu$^\ddag$}
\orcid{1234-5678-9012}
\affiliation{%
  \institution{
  $\dag$Dept. of Comp. Sci. \& Tech., TNList Lab, State Key Lab for Intell. Tech. \& Sys., CBICR Center, Tsinghua University\\
  $\ddag$School of Software, TNList Lab, State Key Lab for Intell. Tech. \& Sys., Tsinghua University, Beijing, 100084, China
  }
}
\email{{chenjian14,luj15}@mails.tsinghua.edu.cn; {dcszj,shixia}@tsinghua.edu.cn; }


\renewcommand{\shortauthors}{Chen, Zhu, Lu, and Liu}

\begin{abstract}
Nested Chinese Restaurant Process (nCRP) topic models are powerful nonparametric Bayesian methods to extract a topic hierarchy from a given text corpus, where the hierarchical structure is automatically determined by the data. Hierarchical Latent Dirichlet Allocation (hLDA) is a popular instance of nCRP topic models. However, hLDA has only been evaluated at small scale, because the existing collapsed Gibbs sampling and instantiated weight variational inference algorithms either are not scalable or sacrifice inference quality with mean-field assumptions. Moreover, an efficient distributed implementation of the data structures, such as  dynamically growing count matrices and trees, is challenging. 

In this paper, we propose a novel partially collapsed Gibbs sampling (PCGS) algorithm, which combines the advantages of collapsed and instantiated weight algorithms to achieve good scalability as well as high model quality. An initialization strategy is presented to further improve the model quality. Finally, we propose an efficient distributed implementation of PCGS through vectorization, pre-processing, and a careful design of the concurrent data structures and communication strategy. 

Empirical studies show that our algorithm is 111 times more efficient than the previous open-source implementation for hLDA, with comparable or even better model quality. Our distributed implementation can extract 1,722 topics from a 131-million-document corpus with 28 billion tokens, which is 4-5 orders of magnitude larger than the previous largest corpus, with 50 machines in 7 hours.
\end{abstract}




\maketitle

\section{Introduction}
Topic models are popular tools in the machine learning toolbox. They extract a set of latent topics from an input text corpus. Each topic is a unigram distribution over words, and the high-probability words often present strong semantic correlation. 
Topic models have been widely used in information retrieval~\cite{wei2006lda}, text analysis~\cite{boyd2007topic,zhu12jmlr}, information visualization~\cite{Wang2016_TVCG_TopicPanorama}, and many other application areas for feature extraction and dimensionality reduction.

However, the traditional topic models, such as Latent Dirichlet Allocation (LDA)~\cite{blei2003latent}, are flat. They do not learn any relationships between topics by assuming that the probabilities of observing all the topics are independent. 
On the other hand, topics are naturally organized in a hierarchy~\cite{paisley2015nested}.
For example, when a topic on ``unsupervised learning'' is observed in a document, it is likely to also observe the more general topics containing the topic, such as ``machine learning'' and ``computer science'' in the same document. By capturing such relationships, hierarchical topic models can achieve deeper understanding and better generalization~\cite{ahmed2013nested,paisley2015nested} of the corpus than the flat models. 



There are many different approaches to learning the topic hierarchy. For example, Google's Rephil~\cite{murphy2012machine} puts a hierarchical noisy-or network on the documents; the super-topic approach learns topics of topics~\cite{li2006pachinko,pujara2012large}; and the nested Chinese Restaurant Process (nCRP)~\cite{blei2010nested, ahmed2013nested, paisley2015nested} approach utilizes the nCRP as a prior on topic hierarchies. 
Hierarchical topic models have been successfully applied to document modeling~\cite{paisley2015nested}, online advertising~\cite{murphy2012machine} and microblog location prediction~\cite{ahmed2013nested}, outperforming flat models. 
Amongst these approaches, the nCRP method has a non-parametric prior on the topic hierarchy structure, which leads to a natural structure learning algorithm with Gibbs sampling, avoiding the slow-mixing Metropolis-Hastings proposals or neural rules~\cite{murphy2012machine}. 
The hierarchical Latent Dirichlet Allocation (hLDA) model is a popular instance of nCRP topic models~\cite{blei2010nested}. 
In hLDA, topics form a tree with an nCRP prior, while each document is assigned with a path from the root topic to a leaf topic, and words in the document are modeled with an admixture of topics on the path. 


However, due to the lack of scalable algorithms and implementations, hLDA has only been evaluated at a small scale, e.g., with thousands of documents and tens of topics~\cite{blei2010nested,wang2009variational}, which limits its wider adoption in real-life applications. The training of topic models can be accelerated via distributed computing, which has been successfully applied to 
LDA to handle hundreds of billions of tokens and millions of topics~\cite{ahmed2012scalable,chen2015warplda}.
Unfortunately, the previous algorithms for hLDA are unsuitable for distributed computing. 
Specifically, the collapsed Gibbs sampler~\cite{blei2010nested} is difficult to parallelize because collapsing the topic distributions breaks the conditional independence between document-wise latent variables; on the other side, the instantiated weight variational inference algorithm~\cite{wang2009variational} has inferior model quality because of label switching and local optimum, as we will analyze in Sec.~\ref{sec:iws}. Moreover, the data structures used by hLDA, such as the dynamically growing count matrices and trees, are much more sophisticated than the data structures for LDA, and their efficient distributed implementations are challenging.

In this paper, we propose a novel partially collapsed Gibbs sampling (PCGS) algorithm, which combines the advantages of the collapsed Gibbs sampler~\cite{blei2010nested} and the instantiated weight variational inference method~\cite{wang2009variational} to achieve a good trade-off between the scalability and the quality of inference. We keep most topic distributions as instantiated to maximize the degree of parallelism; while we integrate out some rapid changing topic distributions to preserve the quality of inference. To further improve the model quality, we propose an initialization strategy. Finally, we present an efficient distributed implementation of PCGS through vectorization, pre-processing, and a careful design of the concurrent data structures and the communication strategy.


We design a set of experiments to extensively examine the model quality of our PCGS algorithm as well as its efficiency and scalability. The experimental results show that our single-thread PCGS is 111 times faster than the previous state-of-the-art implementation, \texttt{hlda-c}~\cite{hlda-c};
and our distributed PCGS can extract 1,722 topics from a 131-million-document corpus with 28-billion tokens, which is 4-5 orders of magnitude larger than the previous largest corpus, with 50 machines in 7 hours. To the best of our knowledge, this is the first time to scale up hLDA for large-scale datasets.


\vspace{-.1cm}
\section{Hierarchical LDA}
We first review nCRP and 
hLDA for learning a topic hierarchy. 

\vspace{-.1cm}
\subsection{Nested Chinese Restaurant Process}\label{sec:ncrp}
Nested Chinese Restaurant process (nCRP)~\cite{blei2010nested} represents a powerful nonparametric Bayesian method to learn a tree structure, whose width and depth are unbounded. 
Suppose there is a truncated tree with $L$ levels,
where each node except the leaves has an infinite number of children. 
An unique ID is assigned to each node, where the root node has the ID 1.
nCRP defines a probability distribution on a series of paths $(\cv_1, \cv_2, \dots)$ on the tree, where each path $\cv_d\in \mathbb N_+^L$  consists of $L$ node IDs from the root to a certain leaf. Given $\cv_1, \dots, \cv_{d-1}$, 
we mark a node $i$ as \emph{visited} if any of the paths passes through it,  
and the next path $\cv_d$ is generated as follows:  (1) let $c_{d1} = 1$; (2) for each level $l=2, \dots, L$, denote $i$ as a shortcut for $c_{d, l-1}$. Assume that there are already $T$ visited nodes, where the children of $i$ are denoted as $t_{i1}, \dots, t_{iK_i}$.
The next node of the path $c_{dl}$ can be generated as 
\begin{align*}
\begin{cases}
p(c_{dl}=t_{ik})=\frac{m_{t_{ik}}}{\gamma_l+m_i}, & k=1, \dots, K_i \\
p(c_{dl}=T+1)=\frac{\gamma_l}{\gamma_l+m_i},&
\end{cases}
\end{align*}
where $m_i := \#\{(d, l) | c_{dl}=i\}$ is the number of times that node $i$ is visited, $\gamma_l$ is a hyper-parameter, and $\#\{\cdot\}$ denotes the cardinality of a set. 
If $c_{dl}=T+1$, the path goes through a child node of $c_{d,l-1}$, which is not visited before, we assign it the ID $T+1$.
We refer this operation as \emph{the generation of a new child}, although in fact it is just visiting a node that is never visited before. The above procedure is denoted as $\cv_d\sim \mbox{nCRP}(\cv_d; \gammav, \cv_{<d})$, where the subscript $<d$ stands for all the possible indices that are smaller than $d$, i.e., $\cv_{<d}=\{\cv_{1}, \dots, \cv_{d-1}\}$. 


Intuitively, nCRP puts a CRP~\cite{teh2011dirichlet} on each parent node, where the probability of visiting each node is proportional to its  previous times of visit. 
Due to this fact, 
we can easily extend the stick-breaking formulation for CRP~\cite{sethuraman1994constructive} to nCRP~\cite{wang2009variational}. 
The generative procedure is:
\begin{itemize}
\item For each node $t\in \mathbb N_+$ on each level $l$, draw $\piv^t\sim \mbox{GEM}(\gamma_l)$, where $\piv^t$ is a distribution over the children of $t$, and $\mbox{GEM}(\cdot)$ is the stick-breaking distribution~\cite{sethuraman1994constructive}. A sample $\piv\sim \mbox{GEM}(\gamma)$ can be obtained as follows:
\begin{itemize}
\item For $i=1, \dots, \infty$, draw $V_i\sim \mbox{Beta}(1,\gamma)$, and let $\pi_i=V_i\prod_{j=1}^{i-1}(1-V_j)$.
\end{itemize}
\item For each path $\cv_d$, let $c_{d1}=1$, and for $l=2, \dots, L$, select a child $k_{dl}\sim \Mult(\piv^{c_{d,l-1}})$, and let the corresponding id  $c_{dl}=t_{c_{d,l-1},k_{dl}}$.
\end{itemize}
The original nCRP is recovered by integrating out $\{\piv^t\}$. 
In the stick-breaking formulation, the probability of visiting each child is explicitly instantiated, and the paths $\{\cv_d\}$ are conditionally independent given the probabilities $\{\piv^t\}$.

\subsection{Hierarchical Latent Dirichlet Allocation}\label{sec:hlda}
Given a corpus of $D$ bag-of-words documents $\Wv=\{\wv_d\}_{d=1}^D$, where each document $\wv_d=\{w_{dn}\}_{n=1}^{N_d}$ has $N_d$ tokens, and each token is represented by its word id $w_{dn}\in \{1, \dots, V\}$ in the vocabulary of $V$ unique words. hLDA is an nCRP-based topic model to learn a topic hierarchy~\cite{blei2010nested}. 
In hLDA, topics form a $L$-level tree, i.e., each tree node $t$ is a topic, and is associated with a distribution over words $\phiv_t \in \Delta^{V-1}$, where $\Delta^{V-1}$ is the $(V-1)$-simplex. Since nodes and topics have one-to-one correspondence, we do not distinguish them in the sequel. 

In hLDA, each document is assigned with a path $\cv_d$, and its words are modeled with a mixture of the topics in $\cv_d$, with the document-specific mixing proportion $\thetav_d$. The generative process for the corpus is:
\begin{itemize}
\item For each node $t$, draw $\phiv_t\sim \Dir(\beta_{l_t} \mathbf{1})$, where $l_t$ is the level of node $t$ and $\mathbf 1 = (1, \dots, 1)$ is an all-one vector;
\item For each document $d$:
\begin{itemize}
\item Draw $\cv_d\sim \mbox{nCRP}(\cv_d; \gammav, \cv_{<d})$;
\item Draw $\thetav_d\sim \Dir(\alphav)$;
\item For each position $n$, draw $z_{dn}\sim \Mult(\thetav_d)$, and draw $w_{dn}\sim \Mult(\phiv_{c_{d, z_{dn}}})$,
\end{itemize}
\end{itemize}
where $\mbox{Dir}(\cdot)$ is the Dirichlet distribution, and $\alphav$ and $\betav$ are Dirichlet 
hyper-parameters.
There are two special cases of hLDA. When the tree degenerates to a chain, hLDA recovers the vanilla LDA with $L$ topics, and when the tree has two levels and the probability of assigning to the first level $\theta_{d1}$ is close to zero, hLDA recovers the Dirichlet Process Mixture Model (DPMM)~\cite{neal2000markov}.
\section{Inference for HLDA}\label{sec:inference}
\begin{figure}[t]
\centering
\subfigure[Parameter server]{
\includegraphics[scale=0.35]{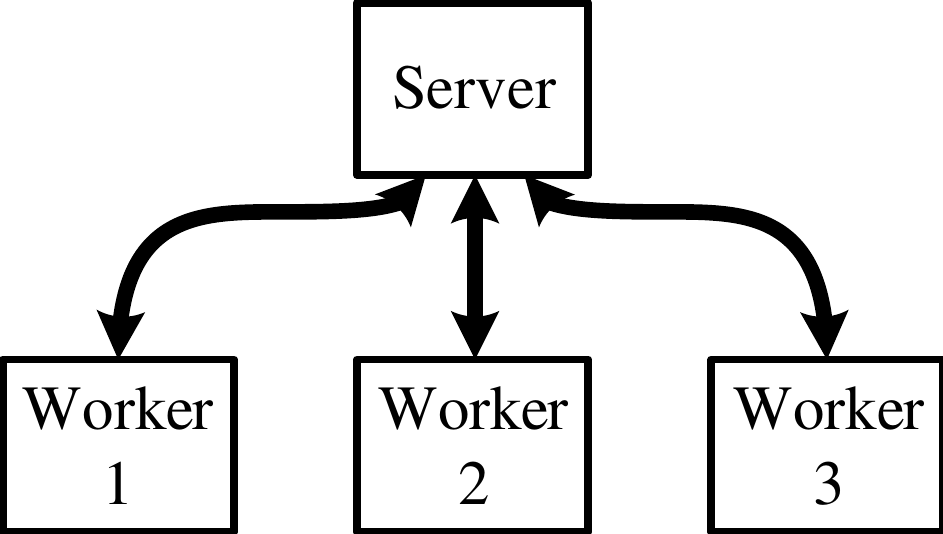}}
\hfill
\subfigure[Bulk synchronous parallel]{
\includegraphics[scale=0.35]{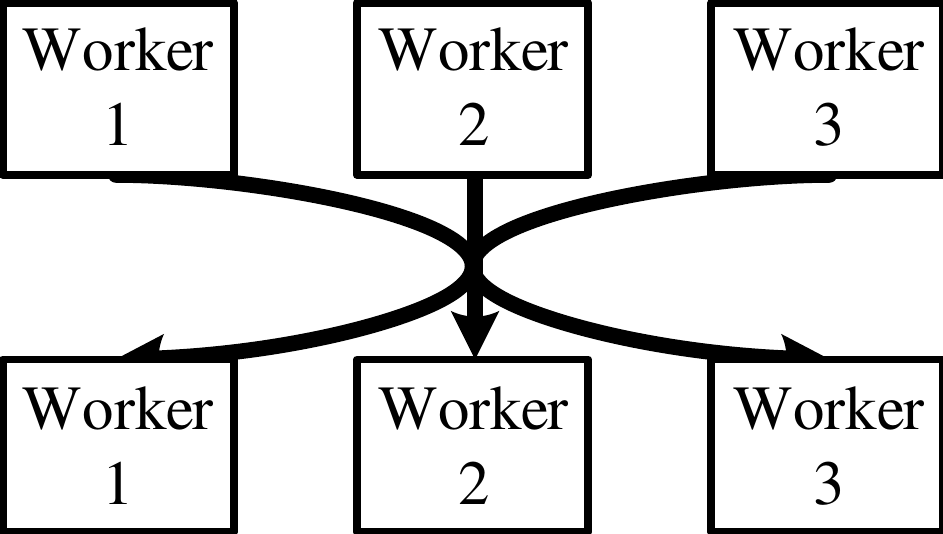}}
\subfigure[Hybrid decentralized and bulk synchronous parallel]{
\includegraphics[scale=0.35]{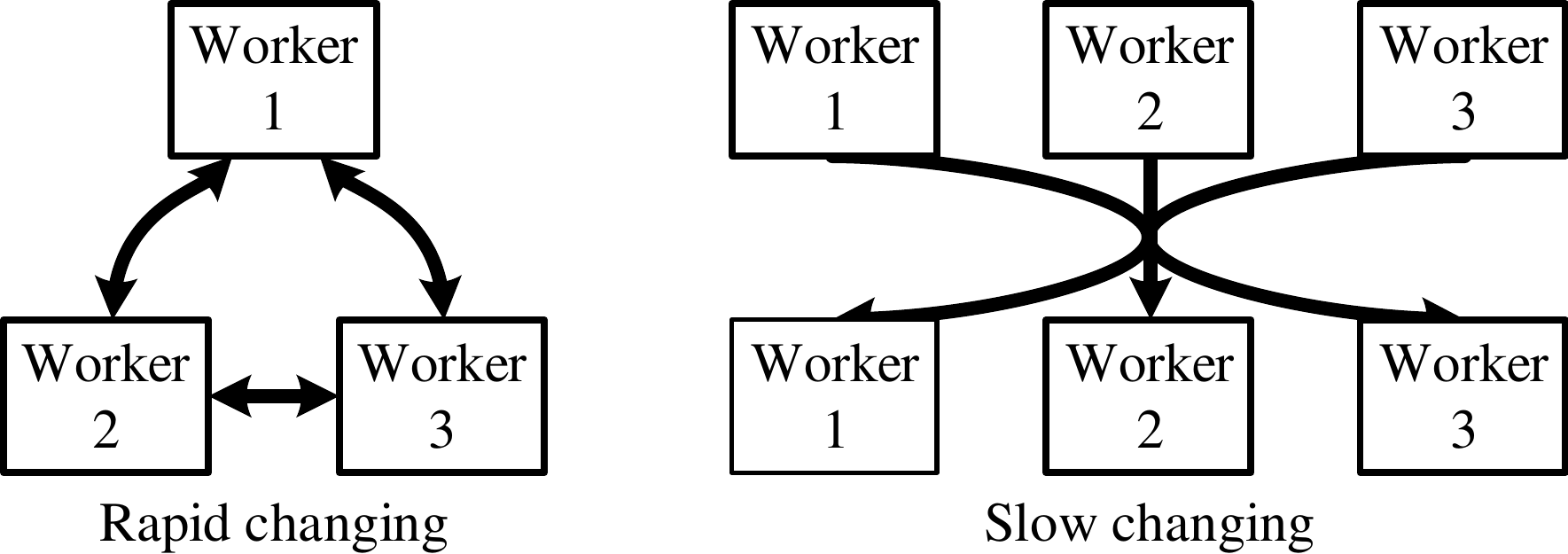}}\vspace{-.3cm}
\caption{Comparison of distributed computing strategies.  }\vspace{-.3cm}
\label{fig:system}
\end{figure}
There are two classes of algorithms for the posterior inference in hLDA---the \emph{collapsed weight algorithm} that integrates out the mixing weights $\piv$ and parameters $\phiv$, and the \emph{instantiated weight algorithm} that explicitly infers these parameters. 
These algorithms present a trade-off between scalability and the quality of inference. 

In this section, we first introduce the collapsed Gibbs sampler~\cite{blei2010nested}, which is not scalable.
To address this problem, we present an instantiated weight block Gibbs sampler, which is based on the same stick-breaking formulation as the variational inference algorithm~\cite{wang2009variational} and has an excellent scalability.
However, it suffers from local optima. 
To tackle this issue, we propose a partially collapsed Gibbs sampler that has good scalability as well as high-quality inference. We also present an initialization strategy to find better local optima.\looseness=-1 



\subsection{Collapsed Gibbs Sampling (CGS)}\label{sec:cgs}
CGS is a collapsed weight algorithm that is based on the nCRP formulation. The generative process of hLDA (Sec.~\ref{sec:hlda}) defines a joint distribution 
$p(\wv, \zv, \thetav, \cv, \phiv)$=
$\prod_{t=1}^{\infty}$ $p(\phiv_t)$ $\prod_{d=1}^D$ $p(\cv_d | \cv_{<d})$ $p(\thetav_d)$
$\prod_{d=1}^D$ $\prod_{n=1}^{N_d}$ $p(z_{dn}|\thetav_d)$  $p(w_{dn} | \phiv_{c_{d, z_{dn}}}).$
Based on the conjugacy between Dirichlet and multinomial, $\thetav$ and $\phiv$ are integrated out to get the collapsed distribution:
\begin{align}
p(\wv, \zv, \cv)=& \prod_{d=1}^D \big[ p(\cv_d | \cv_{<d}) \frac{B(\Cv_d+\alphav)}{B(\alphav)} \big] \prod_{t=1}^\infty \frac{B(\Cv_t+\beta_{l_t})}{B(\beta_{l_t})},\label{eqn:cgs-marginal}
\end{align} 
where $\Cv_d = (C_{d1}, \dots, C_{dL})$ are the document-level counts, i.e., $C_{dl}=\#\{n | z_{dn}=l\}$, $\Cv_t$ are the topic-word counts, i.e., $C_{tv}=\#\{(d, n) | w_{dn}=v \wedge c_{d, z_{dn}}=t\}$, and $B(\cdot)$ is the multivariate beta function, $B(\alphav)=\prod_k \Gamma(\alpha_k) / \Gamma(\sum_k \alpha_k)$.

CGS alternatively samples $\zv$ and $\cv$ 
from their conditional distributions: $p(z_{dn}=l | w_{dn}=v, \wv, \zv_{\neg dn}, \cv) \propto (C_{dl}^{\neg dn}+\alpha_l)\frac{C_{c_{dl},v}^{\neg dn}+\beta_l}{C_{c_{dl}}^{\neg dn}+V\beta_l}$, and $p(\cv_d = \cv | \wv, \zv, \cv_{\neg d}) \propto \mbox{nCRP}(\cv; \gammav, \cv_{\neg d}) \prod_{l=1}^L f_C(d, c_{l})$,
where $\neg d$ represents excluding document $d$, and $\neg dn$ means excluding the token $(d, n)$, e.g., $C_{tv}^{\neg d} = \#\{(d^\prime, n) | w_{d^\prime n}=v \wedge c_{d^\prime,z_{d^\prime n}}=t \wedge d^\prime\ne d\}$. 
Furthermore, the superscript $^d$ denotes only considering document $d$, e.g., $C_{tv}^d = \#\{n | w_{dn}=v \wedge c_{d,z_{dn}}=t\}$. Finally, $C_t = \sum_{v=1}^V C_{tv}$ and 
\begin{align}
f_C(d, t) = \frac{B(\Cv_t^{\neg d} + \Cv_t^{d}+\beta_{l_t})}{B(\Cv_t^{\neg d}+\beta_{l_t})}.\label{eqn:fc}
\end{align}


A straightforward approach for parallelizing CGS is to let each thread work on a disjoint set of documents and synchronize the counts $\{\Cv_t\}$ between threads and machines. 
One possible solution for the  synchronization is the parameter server~\cite{ahmed2012scalable}, which maintains a \emph{local copy} of $\{\Cv_t\}$ on each worker machine as well as on a parameter server. Each machine periodically synchronizes its local copy with the parameter server by pushing its change to the server and fetching the latest parameters. Multiple worker threads read and write the local copy concurrently, as illustrated in Fig.~\ref{fig:system}(a). 

However, this approach has several disadvantages, limiting its scalability and efficiency. Firstly, due to the limited network bandwidth, the period of synchronization can be relatively long, e.g., minutes. While this is acceptable for LDA, the stale state can potentially harm the quality of inference for hLDA, which is much more sensitive to local optima, as we will analyze in Sec.~\ref{sec:iws}. Secondly, the local copy of $\{\Cv_t\}$ needs to support concurrent reads, updates, and \emph{resizes} from the worker threads, which is notoriously difficult and much more expensive than a serial version~\cite{dechev2006lock}. Finally, even in a serial setting, the computational cost of CGS is high because $f_C(d, t)$ involves the computation of the multivariate beta function, which is computed with gamma functions (see the appendix), that are much more expensive to compute than simple arithmetic.

\subsection{Block Gibbs Sampling}\label{sec:iws}
To address the scalability and efficiency issues of CGS, we begin with a block Gibbs sampler (BGS), which is an instantiated weight algorithm that is based on the same model formulation of the variational inference algorithm~\cite{wang2009variational}, but the per-iteration time complexity is made lower by replacing expectation with sampling.

The BGS is based on the stick-breaking formulation of nCRP (defined in Sec.~\ref{sec:ncrp}), which defines a joint distribution $p(\wv, \zv, \thetav, \cv, \phiv, \piv)$. Integrating out $\thetav$, BGS samples in the posterior distribution of $(\zv, \cv, \phiv, \piv)$ by alternatively sampling $\zv$, $\cv$, $\phiv$ and $\piv$ given the others. The resultant updates are as follows:

\textbf{Sample $\zv$:} for each token, draw a level assignment from the distribution $p(z_{dn}=l | w_{dn}=v, \wv, \cv, \phiv, \piv) 
\propto (C_{dl}^{\neg dn}+\alpha_l) \phi_{c_{dl}, v}.$

\textbf{Sample $\cv$:} for each document, sample a path from the conditional distribution
$p(\cv_d=\cv | \wv, \zv, \cv_{\neg d}, \phiv, \piv)\propto \prod_{l=2}^L \pi_{c_{l-1}\rightarrow c_l} \prod_{l=1}^L$ $f_I(d, c_l),$
where $\pi_{i\rightarrow j}$ is the probability of going from node $i$ to its child $j$, i.e., $\pi^i_k = \pi_{i\rightarrow t_{ik}}$, 
and \begin{align}
f_I(d, t) = \prod_{v=1}^V \phi_{tv}^{C_{tv}^d}.\label{eqn:fi}
\end{align}

\textbf{Sample $\piv$:} 
Draw the stick-breaking weights $\pi^t_k = V^t_k\sum_{j=1}^{k-1}(1-V^t_j)$, where $V^t_k\sim \mbox{Beta}(1+m^t_k, \gamma+m^t_{>k})$, $m^t_k$ is the number of times that a path go through $t$ and its $k$-th child, and $m^t_{>k}=\sum_{j=k+1}^{\infty} m^t_j$.

\textbf{Sample $\phiv$:} Draw the topic distribution $p(\phiv_t) = \Dir(\beta_{l_t} + \Cv_t)$.

A subtlety here is on sampling $\piv$ and $\phiv$, where $\piv^t$ is infinite-dimensional, and there are infinite $\phiv_t$'s. We approximate the sampling by truncating $\piv^t$ to be finite-dimensional, i.e., there are finite children for each node, so that the whole tree has a finite number of nodes. This approximation can be avoided with slice sampling~\cite{ge2015distributed}, but truncation is not the main reason affecting the model quality, as there are some more severe issues as we shall see soon. 

Due to conditional independence, the document-specific random variables $\zv$ and $\cv$ can be sampled in parallel for each document. This fits in a bulk-synchronous parallel (BSP) pattern. In each iteration, $\piv$ and $\phiv$ are sampled and broadcast-ed to each worker, and then the workers sample $\zv$ and $\cv$ without any communication, as illustrated in Fig.~\ref{fig:system}(b). BSP has been successfully adopted for LDA~\cite{zhai2012mr,zaheer2016exponential,chen2015warplda} and achieved superior throughput than the asynchronous version (e.g., parameter server) for being lock free~\cite{zaheer2016exponential}. Moreover, the BGS update for $\cv$ is also cheaper than that of CGS because the computation of $f_I(d, t)$ only involves the logarithm of $\phiv$, which remains invariant during the sampling of $\zv$ and $\cv$, and therefore, can be pre-processed.

Unfortunately, while BSP works well for LDA, its quality of inference is unsatisfactory for hLDA in practice. We provide a number of explanations for this phenomenon:

\noindent\textbf{$\vect C_t$ is not slow-changing.} 
BSP works well for LDA because the topic-word count $\vect C_t$ is slow-changing. Therefore, a stale $\vect C_t$ is close to the fresh version, and the result is not much affected. However, this assumption is not true for hLDA because a topic can have very few documents assigned to it. 
For instance, if the sampler assigns a document to a topic that do not have any assigned documents, the topic-word count $\vect C_t$ for that topic will suddenly change from a zero vector to non-zero, which differs significantly with its stale version. \looseness=-1

\noindent \textbf{Label switching.}
If two different workers generate two new topics, it is not clear whether they are the same one. For example, in a certain iteration, documents $d_1$ and $d_2$ should be assigned to two different new topics $t_1$ and $t_2$. But in BGS, two different workers may decide to assign $d_1$ and $d_2$ to a \emph{same} topic $t$, because both workers do not know the changes made by the other worker, and just regard $t$ as a topic that no document is assigned to it. As the result, instead of learning two different topics $t_1$ and $t_2$, BGS learns one topic $t$ that is a combination of $t_1$ and $t_2$. 
For flat models such as DPMM or hierarchial Dirichlet process (HDP)~\cite{teh2004sharing}, label switching is sometimes (approximately) resolved by running an algorithm that matches the new topics from different workers~\cite{campbell2015streaming}. However, it is not clear how to match topics on trees.

\noindent \textbf{Local optima.}
In flat models, even when label switching happens, e.g., two topics $t_1$ and $t_2$ are mixed as one topic, the algorithm may gradually separate them by generating a new topic and assigning the documents that should belong to $t_2$ to the new topic~\cite{yang2016distributing}. However, these moves are more difficult for hLDA because it is more sensitive to local optima. For instance, if two topics $t_1$ and $t_2$ are incorrectly mixed as one topic $t$, and $t$ \emph{has a sub-tree}. To correctly separate $t_1$ and $t_2$, the sampler needs to create a new brother of $t$, and move some decedents of $t$ to its brother. These operations can hardly be achieved with local moves. Wang and Blei \cite{wang2009variational} attempted to make this kind of moves by split-and-merge operations, whose time complexity is typically quadratic with the number of topics, and does not scale to a large number of topics.


\subsection{Partially Collapsed Gibbs Sampling}\label{sec:pcgs}
It can be seen from the aforementioned discussion that  there is a trade-off between scalability and the quality of inference. CGS learns good models but is not scalable, while BGS is very scalable but sacrifices the quality of inference. 
To combine their advantages, we propose a partially collapsed Gibbs sampler (PCGS).\looseness=-1

Intuitively, if a topic $t$ has lots of assigned documents, its count $\vect C_t$ changes slowly. Based on this insight, we categorize the topics as slow changing topics (SCTs) $\mathcal I$ and rapid changing topics (RCTs) $\mathcal C$, such that the number of SCTs dominates. Then, we perform CGS for the RCTs $\mathcal C$ and BGS for the SCTs $\mathcal I$. The quality of inference is not greatly affected because we perform CGS for the RCTs, and the scalability and efficiency is good because for most topics we perform the scalable and efficient BGS. We define a topic $t$ to be rapid changing if it is assigned to less than $M$ (a user-defined constant) documents, and slow changing otherwise. 

Formally, let $\phiv_{\mathcal C} = \{\phiv_t | t\in \mathcal C\}$, $\phiv_{\mathcal I} = \{\phiv_t | t\in \mathcal I\}$, $\Cv_{\mathcal C}=\{\Cv_t | t\in \mathcal C\}$ and $\Cv_{\mathcal I}=\{\Cv_t | t\in \mathcal I\}$, we derive the following Gibbs sampling updates, where the details can be found in the appendix:

\textbf{Sample $\zv$:} draw the level assignment for each token from 
$
p(z_{dn}=l | w_{dn}=v, \wv, \zv_{\neg dn}, \cv, \phiv_{\mathcal I})
$
$\propto(C_{dl}^{\neg dn}+\alpha_l) \begin{cases}
 \phi_{c_{dl}, v} & c_{dl} \in \mathcal I, \\
 \frac{C_{c_{dl},v}^{\neg dn}+\beta_{l_t}}{C_{c_{dl}}^{\neg dn}+V\beta_{l_t}} & c_{dl} \in \mathcal C.
\end{cases}
$

\textbf{Sample $\cv$:} sample the path from  
$p(\cv_d=\cv | \wv, \zv, \cv_{\neg d}, \phiv_{\mathcal I})$
\begin{align}
\propto \mbox{nCRP}(\cv; \gammav, \cv_{\neg d}) \prod_{l=1}^L 
\begin{cases}
f_I(d, c_{l}) & c_{l}\in \mathcal I,\\
f_C(d, c_l) & c_{l}\in \mathcal C.
\end{cases}\label{eqn:pcgs-c}
\end{align}

\textbf{Sample $\phiv^\mathcal I$:} For $t\in \mathcal I$, draw $\phiv_t \sim \Dir(\beta_{l_t} + \Cv_t)$.

These PCGS updates just combine the update rules of CGS and BGS, which utilizes CGS rule for $t\in \mathcal C$ and BGS rule for $t\in \mathcal I$.

Since the document visit counts $\mv$ (defined in Sec.~\ref{sec:ncrp}) only requires $O(T)$  space, where $T$ is the number of topics, it is cheap to synchronize. We keep the entire tree weight $\piv$ collapsed out, and periodically synchronize the counts across machines. PCGS creates new topics in the same way as CGS. Therefore, PCGS does not require truncation and has the correct stationary distribution.

For distributed computing, PCGS performs asynchronous updates for the rapid-changing $(\Cv_\mathcal C, \mv)$ and perform BSP updates for the slow-changing counts $\Cv_\mathcal I$, as illustrated in Fig.~\ref{fig:system}(c). Since there are few rapid-changing topics, the amount of asynchronous updates of PCGS is much smaller than that of CGS, which needs to update all the counts $(\{\Cv_t\}, \mv)$ asynchronously. Thanks to the small size of asynchronous updates, network bandwidth is not a bottleneck for PCGS, and PCGS can update the counts more frequently than CGS. In the sequel, the PCGS counts are more fresh than CGS in distributed setting. Because the number of slow-changing topics dominates, PCGS enjoys similar scalability and efficiency as BGS.

\subsection{Initialization Strategy}\label{sec:initialization}
hLDA is sensitive to local optima, so a proper initialization is crucial for obtaining good results. We adopt the progressive online initialization strategy~\cite{blei2010nested,wang2009variational}, which begins with an empty collection of documents, and gradually adds documents by inferring the posterior of document-specific variables $(\cv_d, \zv_d)$ given all the previously observed documents. The documents are organized into mini-batches, and $\phiv_\mathcal I$ is  sampled per mini-batch.

To further improve the model quality, we noticed that all the aforementioned algorithms update $\cv$ and $\zv$ while keeping the other fixed, 
which can severely trap the sampler in local optima. For example, after a document $d$ is assigned to a certain path $\cv_d$, its words are assigned to the levels $\zv_d$ \emph{of the current path}. In the next iteration, even if there is another path $\cv^\prime_d$ such that $p(\wv_d | \cv^\prime_d)$ is larger than $p(\wv_d| \cv_d)$, $p(\wv_d | \cv^\prime_d, \zv_d)$ is not likely to be larger than $p(\wv_d | \cv_d, \zv_d)$ because $\zv_d$ is already optimized for $\cv_d$. In this case, the path assignments $\cv_d$ will be quickly trapped in a local optimum even if there are better path assignments. We also noticed that similar as in multinomial mixture  models~\cite{rigouste2007inference}, the sampling of $\cv_d$ is almost deterministic, because $\log p(\wv_d | \cv_d, \zv_d)$ is a sum of log-likelihoods over all the words, and can differ by hundreds for different $\cv_d$'s. Therefore, it is difficult for a sampler to jump out of the local trap simply by its randomness.

We propose a remedy for this problem by sampling $\cv$ from $p(\cv | \wv)$ directly instead of from $p(\cv | \zv, \wv)$ (Eq.~\ref{eqn:pcgs-c}) for the first $I$ iterations. In other words, we integrate $\zv$ out. In the first $I$ iterations, the sampler focuses on finding the optimal assignment $\cv$ for each document. Afterwards, the algorithm samples $p(\cv | \zv, \wv)$ to refine the model. Unfortunately, $p(\cv | \wv)=\sum_{\zv} p(\cv|\zv, \wv) p(\zv)$ has no closed-form representation. We approximate it with Monte-Carlo integration
$p(\cv | \wv)\approx \frac{1}{S}
\sum_{\zv_s\sim p(\zv)} p(\cv | \zv_s, \wv),$
where $S$ is the number of samples, and $p(\zv)=\int_{\thetav} p(\zv|\thetav)p(\thetav) d\thetav$ is a Polya distribution which is approximated with a uniform discrete distribution over levels. 



\section{System Implementation}

\begin{figure}[t]
\centering
\includegraphics[width=\linewidth]{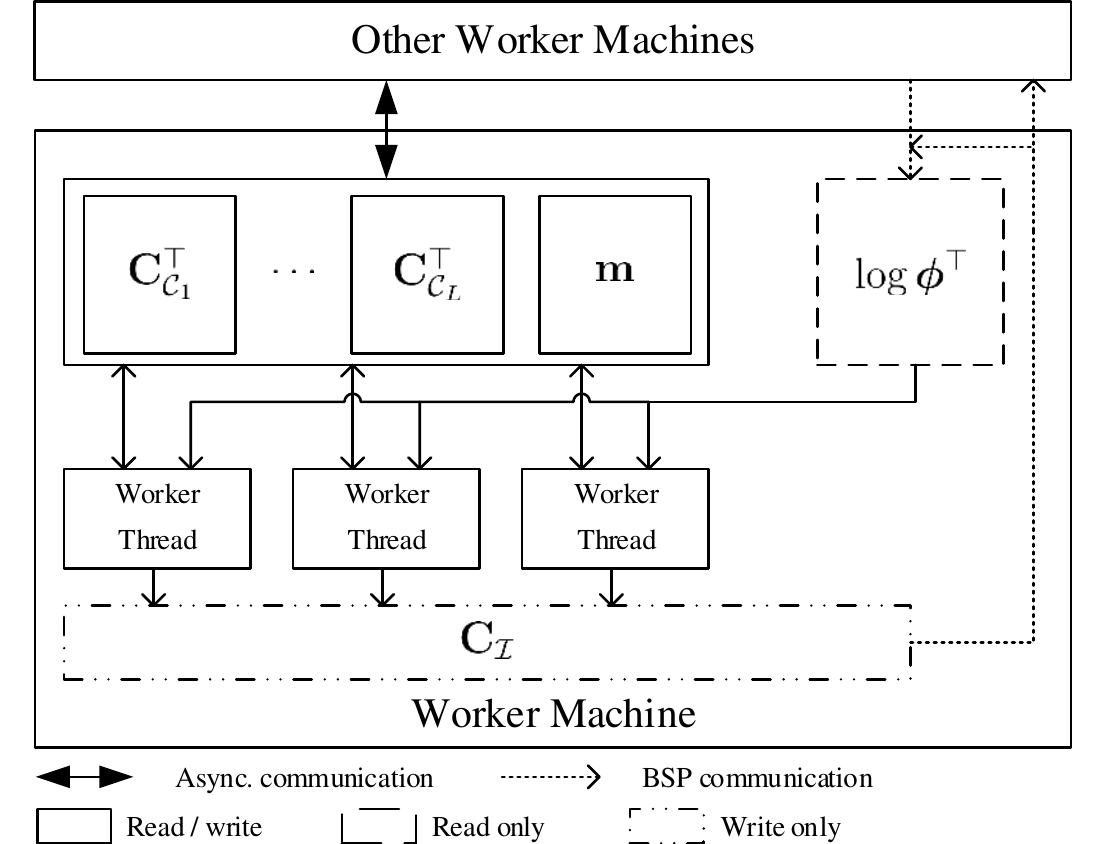}\vspace{-.2cm}
\caption{An overview of the distributed inference  system.\label{fig:system-overview}}\vspace{-.2cm}
\end{figure}

Our distributed training system for hLDA consists of machine-level and thread-level parallelism, as shown in Fig.~\ref{fig:system-overview}. On the machine level, we use MPI to synchronize the tree structure and the counts $(\Cv_{\mathcal I}, \Cv_{\mathcal C}, \mv)$ across machines; and on the thread-level, a number of threads concurrently read and update the local counts. 

In order to implement the system efficiently, several challenges must be addressed. Firstly, for each worker thread, the data layout and algorithm should be organized in a vectorization-friendly way for efficient memory access and computation. Moreover, expensive computation such as logarithms should be avoided as much as possible. Secondly, for efficient multi-thread parallelism, the shared data structure should be lock-free. Finally, the communication strategy need to be chosen carefully to minimize communication overhead and maximize the freshness of the counts. We now present solutions to address these challenges.



\subsection{Vectorization and Pre-processing}\label{sec:data-layout}





We first discuss how to organize the computation for vectorization and pre-processing. 
The most time-consuming part of PCGS is the sampling of $\cv$ according to Eq.~(\ref{eqn:pcgs-c}), or more concretely, computing $f_I(d, t)$ for each $t\in \mathcal I$ and $f_C(d, t)$ for each $t\in \mathcal C$. 
Because both $f_I(d, t)$ and $f_C(d, t)$ are very close to zero, we compute their logarithms. Rewrite Eq.~(\ref{eqn:fi}):
\begin{align}
\log f_I(d, t) = \log \prod_{v=1}^V (\phi_{tv})^{C_{tv}^d}
= \sum_{v\in \mathcal W_{d,l_t}} \log \phi_{tv},\label{eqn:efi}
\end{align}
where $\mathcal W_{dl}$ is the set of all tokens in document $d$ that are assigned to level $l$, which can be computed by bucket sorting $\zv_{dn}$'s along with their positions $n$'s. 

Eq.~(\ref{eqn:efi}) can be vectorized as $\log \fv_I(d, l) = \sum_{v\in \mathcal W_{dl}} (\log \phiv^\top)_{v,\mathcal I_l}$, where $\mathcal I_l \subset \mathcal I$ is the subset of topic ids on level $l$, and $\fv_I(d, l)$ is the vector of $f_I(d, t)$'s for all topics in $\mathcal I_l$. 
The matrix $\log \phiv^\top$ is the transpose of the element-wise logarithm of $\phiv$, which is pre-processed to avoid the computation of logarithm. We store the matrix in row-major order, so that accessing some topics $\mathcal I_l$ for a certain word $v$ is continuous as long as the IDs $\mathcal I_l$ are continuous for each level, which is easy to achieve by sorting the topic ids according to their levels, since $\mathcal I$ do not change when sampling $\cv$. Therefore, computing $\log \fv_I(d, l)$ is fully vectorized by just adding the slices of the matrix $\log \phiv^\top$ indexed by $\mathcal W_{dl}$ and $\mathcal I_l$.


Similarly, Eq.~(\ref{eqn:fc}) can be rewritten as:
\begin{align}
\log f_C(d, t)
=&\sum_{(v, o)\in \mathcal W^\prime_{d,l_t}} 
\log (C_{tv}^{\neg d} + o + \beta_{l_t}) + h_t,\label{eqn:efc}
\end{align}
where we convert the logarithm of multivariate beta function as the sum of logarithms (the derivation details can be found in the appendix).  The term $h_t=\log\Gamma(C_t^{\neg d}+V\beta_{l_t}) - \log\Gamma(C_t+V\beta_{l_t})$, and in $\mathcal W_{dl}^\prime$ we assign each token with an \emph{offset} indicating which time does this word appear, e.g., if a word $v$ is in $\wv_d$ for three times, we put $(v, 0)$, $(v, 1)$ and $(v, 2)$ into $\mathcal W_{dl}^\prime$.

Again, we vectorize the computation of $f_C(d, t)$ by computing $\log \fv_C(d, l)$ for each $l$, which is the vector of $\log f_C(d, t)$'s for $t\in \mathcal C_l$. $\mathcal C_l\subset \mathcal C$ is the subset of topic ids on level $l$, that can change during the sampling of $\cv$ due to the birth and death of topics. 
For efficient vectorization, the counts need to be stored such that $C_{tv}^{\neg d}$ is continuous for all $t\in \mathcal C_l$. To achieve this, we store separate count matrices for each level. For level $l$, $\Cv_{\mathcal C_l}^{\top}$ is stored, which is made by concatenating all the columns $t\in \mathcal C_l$ of $\Cv_{t}^\top$. When a new topic on level $l$ is created, we append it to $\Cv_{\mathcal C_l}^{\top}$ as the rightmost column. The removal of columns is deferred after the sampling of $\cv$ finishes, and the result will not be affected since the dead topics correspond to zero columns in $\Cv_{\mathcal C_l}^{\top}$.


Unlike computing $f_I(d, t)$, the logarithm in Eq.~(\ref{eqn:efc}) cannot be pre-processed since the count $C_{tv}^{\neg d}$ changes during the sampling of $\cv$. Therefore, $f_C(d, t)$ is much more expensive to compute than $f_I(d, t)$, supporting our argument on the inefficiency of CGS in Sec.~\ref{sec:cgs}. Fortunately, for PCGS, the computation of logarithm is avoided as much as possible by keeping $\mathcal C$ a small set. To further accelerate the computation, we use the SIMD enabled Intel VML 
library for logarithms.

\subsection{Concurrent Data Structures}
In our system, the collapsed count matrices $\Cv^{\top}_{\mathcal C}$ are concurrently read and updated by the worker threads, and the number of columns (topics) can grow over time. Since there are a lot of reads, the matrix must be read efficiently, i.e., lock free. Meanwhile, the consistency can be relaxed since a small deviation of the counts will not affect the result much. Therefore, we only ask the matrices to have \emph{eventual consistency}, i.e., the values of the matrices should be eventually correct if no new updates are given. We adopt atomic writes to preserve eventual consistency, while the reads are relaxed as non-atomic operations, to maximize the reading performance.


The dynamic number of columns makes the implementation challenging. The straightforward implementation for growing the matrix involves allocating a new memory region, copying the original content to the new memory, and deallocating the original memory. However, this implementation cannot achieve eventual consistency because the updates during copying will not be incorporated. 

Inspired by the lock-free design of a concurrent vector~\cite{dechev2006lock}, which is a one-dimensional version of our matrix, we provide an efficient implementation of the concurrent matrix. Internally, it holds a list of \emph{matrix blocks}, where the $i$-th matrix block has the size $R\times 2^{c+i-1}$, while $c$ is a constant. The first matrix block represents the $[0, 2^c)$-th columns of the original matrix, the second matrix block represents the $[2^c, 3\times 2^c)$-th columns of the original matrix, and so on. 
If there is a growing request that exceeds the current capacity, we allocates the next matrix block on the list.  For every reading and updating request, the requested $(r, c)$ coordinate is converted to the $(r, b, c^\prime)$ coordinate, where $b$ is the index of the matrix block on the list and $c^\prime$ is the column index within the matrix block. The coordinate conversion can be achieved with a BSR instruction in modern x86 systems~\cite{dechev2006lock}. 
Finally, to improve the locality, we defragment after each PCGS iteration, i.e., deallocating all the matrix blocks and concatenating their content to form a single larger matrix block.

\subsection{Communication}
For PCGS, we need to synchronize the instantiated count $\Cv_\mathcal I$ across machines once per PCGS iteration, and the collapsed counts ($\Cv_{\mathcal C}$, $\mv$) as frequently as possible. We now present an implementation of the synchronization by MPI. 

Firstly, we synchronize $\Cv_\mathcal I$ by the \texttt{MPI\_Allreduce} operation. There are many approaches to synchronizing $\Cv_\mathcal C$ and $\mv$. One possible solution is the parameter server as shown in Fig.~\ref{fig:system}(a). However, while parameter server typically assumes the amount of communication is high and the network bandwidth is the bottleneck, the amount of our PCGS communication is low and our main focus is on the latency, which determines how fresh the count is. The parameter server first merges the changes from individual worker machines at the server, and then pushes the new state to the workers. While the merging decreases the amount of communication, it increases the latency by sending the change through the server. 

To optimize the latency, we design a decentralized communication strategy, in which all the worker nodes directly send their changes to all the other nodes, as illustrated in Fig.~\ref{fig:system}(c). There is a synchronization thread on each worker machine with a \texttt{to\_send} buffer, a \texttt{sending} buffer and a \texttt{receiving} buffer. The worker threads write their changes to the \texttt{to\_send} buffer, and the synchronization threads periodically exchange the content in the \texttt{to\_send} buffer across machines, as follows: (1) Atomically exchange the \texttt{to\_send} buffer and \texttt{sending} buffer, clear the new \texttt{to\_send} buffer; (2) Gather the content of all \texttt{sending} buffers to the \texttt{receiving} buffer, by a \texttt{MPI\_Allgatherv} operation; (3) Merge all the changes in the \texttt{receiving} buffer to the local copy of collapsed counts $\Cv_\mathcal C$ and $\mv$.

\begin{table}[t]
\caption{Statistics of the datasets. \label{tbl:datasets}}\vspace{-.1cm}
\centering
\begin{tabular}{llll}
\hline
Dataset & $D$ & \# tokens & $V$  \\
\hline
NYTimes (subset) & $3\times 10^3$ & $7.23\times 10^5$ & 101635 \\
NIPS & $1.5\times 10^3$ & $1.93\times 10^6$ & 12375 \\
NYTimes & $2.93\times 10^5$ & $9.7\times 10^7$ & 101635 \\
PubMed & $8.2\times 10^6$ & $7.38\times 10^8$ & 141043 \\
ClueWeb12 (small) & $1.5\times 10^7$ & $5.6\times 10^9$ & 100000 \\
ClueWeb12 (large) & $1.31\times 10^8$ & $2.8\times 10^{10}$ & 100000 \\
\hline
\end{tabular}\vspace{-.2cm}
\end{table}

\begin{figure*}[t]
\centering
\begin{minipage}[c]{0.65\linewidth}
\includegraphics[width=0.43\linewidth]{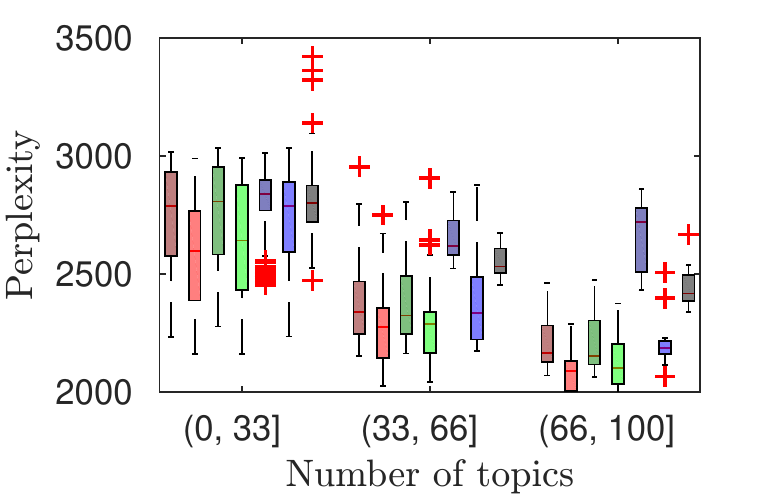}
\includegraphics[width=0.56\linewidth]{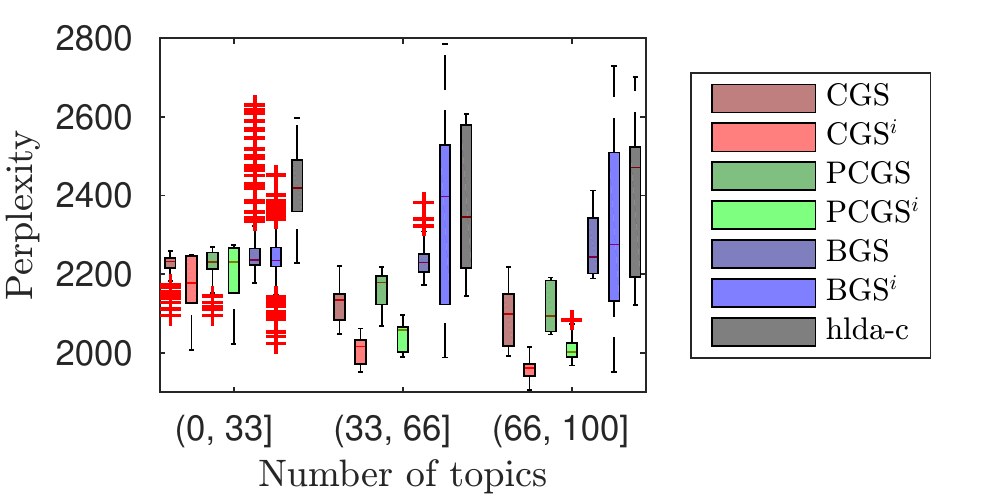}\vspace{-.2cm}
\caption{Comparison of inference quality. The superscript $^{i}$ denotes for our initialization strategy. Left: NYTimes (subset) dataset; Right: NIPS dataset. \label{fig:quality}}
\end{minipage}
\hfill
\begin{minipage}[c]{0.29\linewidth}%
\captionof{table}[foo]{Running time comparison of single-thread implementations. \label{tbl:running-time}}
\begin{tabular}{ll}
\hline
Implementation & Time (s) \\
\hline
hlda-c & 4200 \\
CGS$^i$ & 87.3 \\
PCGS$^i$ & 37.7 \\
BGS$^i$ & 29.5 \\
\hline
\end{tabular}
\end{minipage}\vspace{-.2cm}
\end{figure*}

\begin{figure*}[t]
\centering
\begin{minipage}[t]{0.33\linewidth}
\includegraphics[width=0.49\linewidth]{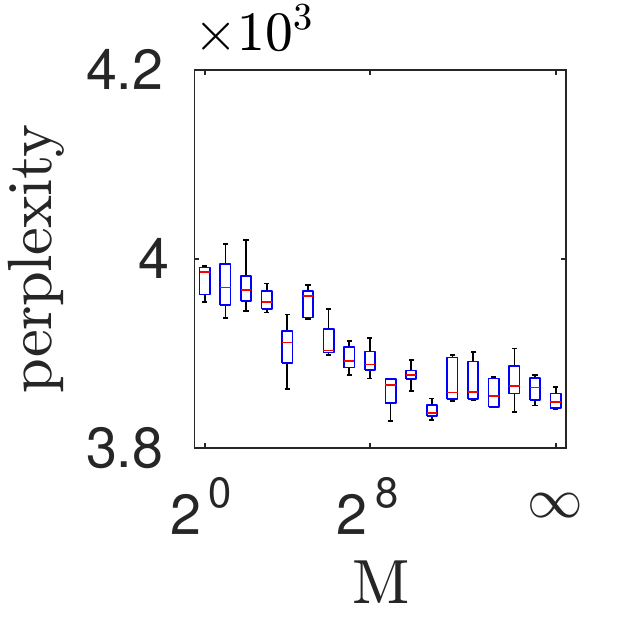}
\includegraphics[width=0.49\linewidth]{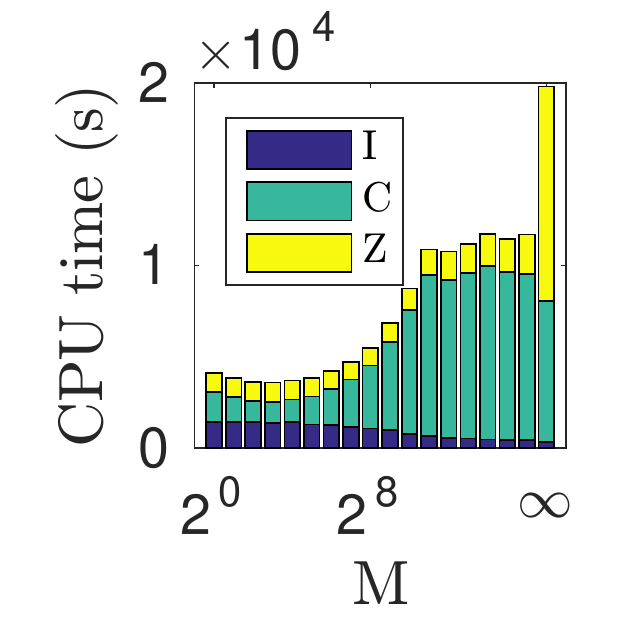}\vspace{-.35cm}
\caption{The impact of $M$. I: computing $f_I$, C: computing $f_C$, Z: sampling $\zv$.
\label{fig:m}}
\end{minipage}
\hfill
\begin{minipage}[t]{0.33\linewidth}
\includegraphics[width=\linewidth]{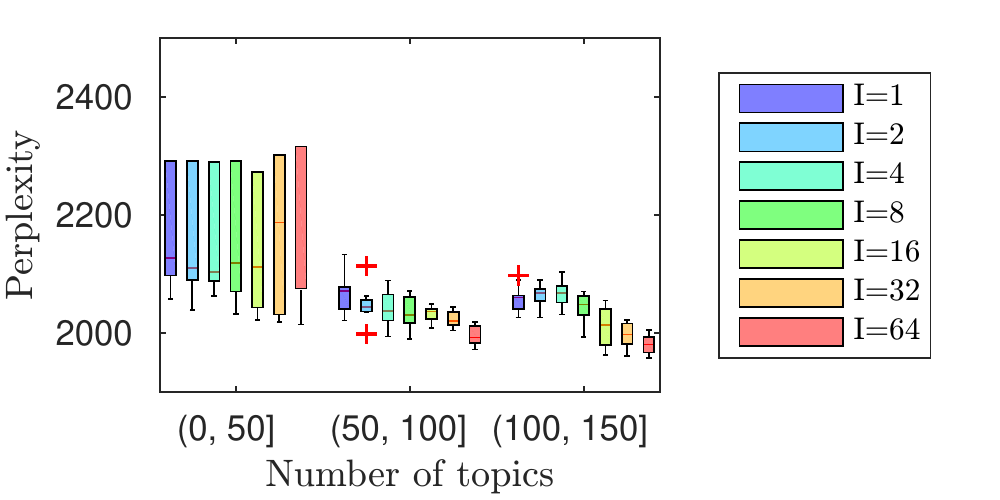}\vspace{-.35cm}
\caption{Impact of $I$ to perplexity.
\label{fig:iters}}
\end{minipage}
\hfill
\begin{minipage}[t]{0.33\linewidth}
\includegraphics[width=0.49\linewidth]{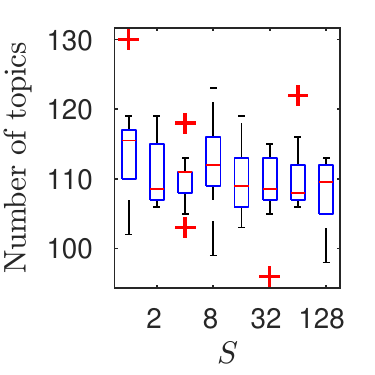}
\includegraphics[width=0.49\linewidth]{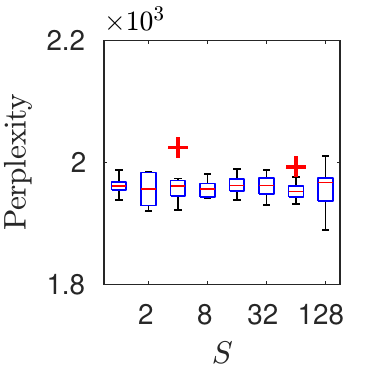}\vspace{-.35cm}
\caption{Impact of the number of Monte-Carlo samples $S$.
\label{fig:num-samples}}
\end{minipage}\vspace{-.3cm}
\end{figure*}


\section{Experiments}
We evaluate the quality, efficiency and scalability of our algorithm and system on several datasets, including NIPS, NYTimes, PubMed from the UCI machine learning repository~\cite{asuncion2007uci}, and two subsets of the ClueWeb12 dataset~\cite{clueweb12} (Table~\ref{tbl:datasets}). The experiments are conducted on the Tianhe-2 supercomputer, which has two 12-core Xeon E5-2692v2 CPUs per node and an InfiniBand network. Our quantitative and qualitative results demonstrate the promise.

We quantitatively compare the quality of the inferred models by predictive log-likelihood using the document completion approach~\cite{wallach2009evaluation}. The corpus is divided as a \emph{training corpus} $\wv^{\mbox{t}}$ and a \emph{testing corpus}, and the testing corpus is further divided as an \emph{observed} corpus $\wv^{\mbox{o}}$, which contains a random half of the tokens for each document in the testing corpus; and a \emph{heldout} corpus $\wv^{\mbox{h}}$ of the other half of the tokens. The predictive log-likelihood is defined as
$p(\wv^{\mbox{h}} | \wv^{\mbox{o}}, \phiv)$,
where the model $\phiv$ is inferred from the training corpus $\wv^{\mbox{t}}$, and is approximated with  a Monte-Carlo integration:
$$
p(\wv^{\mbox{h}} | \wv^{\mbox{o}}, \phiv) \approx
\prod_{d=1}^D \frac{1}{S}\sum_{s=1}^S \prod_{n=1}^{L_d} \sum_{l=1}^L
p(\wv^{\mbox{h}}_{dn}, \zv^{\mbox{h}}_{dn}=l | \cv_d^{(s)}, \thetav_d^{(s)}, \phiv ),
$$
where $\cv_d^{(s)}$ and $\thetav_d^{(s)}$ are the samples from the posterior distribution $p(\cv_d^{(s)}, \thetav_d^{(s)} | \wv^{\mbox{o}}, \phiv)$, which can be obtained with Gibbs sampling, and $S$ is the number of samples. Finally, we convert predictive log-likelihood to predictive perplexity
$$\mbox{perplexity} = \exp(-\mbox{log likelihood} / \mbox{number of tokens}),$$ 
where a lower perplexity score indicates a better model. 

\subsection{Quality of Inference}

We first compare 
the model inferred by CGS and our proposed BGS and PCGS, and examine the effect of the initialization strategy (Sec.~\ref{sec:initialization}). 
We also include 
a comparison with the open source implementation for hLDA, \texttt{hlda-c}, which is a CGS algorithm but has a stick-breaking prior on $\thetav$ instead of a Dirichlet prior~\cite{blei2010nested}.

Unlike parametric models, where the number of topics is fixed, nonparametric models such as hLDA produce different numbers of topics for different runs and various inference algorithms, even with the same hyper-parameter setting. It is not fair to directly compare the perplexity of two models with different numbers of topics. For a fair comparison, we choose a rich set of hyper-parameter configurations, run the algorithms for all these configurations, and plot the perplexity against the number of topics as in Fig.~\ref{fig:quality}. 
In this experiment, we train a 4-layer model (i.e., $L=4$) on the NYTimes (subset) dataset and the NIPS dataset, and $\betav=(\beta_0, 0.5\beta_0, 0.25\beta_0, 0.25\beta_0)$, where $\beta_0$ is chosen from $\{e^{-4.0}, e^{-3.5}, \dots,  e^{2.0}\}$, $\gamma$ is chosen from $\{e^{-6.0}, e^{-5.5}, \dots, e^{0.0}\}$, and $\alphav=0.2\times \mathbf 1$.

By comparing the perplexity produced by different algorithms, we have a number of observations:
\begin{itemize}
\item CGS and PCGS have similar quality, while BGS has worse results.
This agrees with our previous analysis (Sec.~\ref{sec:iws}) that BGS suffers from label switching and local optimum.
\item Our initialization strategy helps obtain better results for both CGS and PCGS.
\item Our result is not worse (actually better) than \texttt{hlda-c}. The discrepancy attributes to the different choice of prior on $\thetav$.
\end{itemize}


\subsection{Efficiency}


We compare the efficiency of our algorithms against  \texttt{hlda-c}. We run the \emph{seria}l version of all the  algorithms for 70 iterations on the NYTimes (subset) dataset while setting $\betav=(1, 0.5, 0.25, 0.1)$, and $\gamma$ is tuned to keep the number of topics around 300. The timing result is shown in Table~\ref{tbl:running-time}. 
Our CGS implementation is 48 times faster than \texttt{hlda-c}.
The significant gain of efficiency attributes to our vectorization and the conversion of the logarithm of gamma function to the sum of logarithms in Sec.~\ref{sec:data-layout} and the appendix. 
PCGS is 2.3 times faster than CGS, and BGS is 1.3 times faster than PCGS. These results match our analysis in Sec.~\ref{sec:data-layout} on that BGS and PCGS are more efficient than CGS. Overall, our PCGS implementation is 111 times faster than \texttt{hlda-c}.

Combining the results on inference quality and efficiency, we find PCGS to be a good trade-off between quality and efficiency by providing the CGS-level quality within BGS-level time consumption.


\subsection{Sensitivity of Parameters}

We now examine the impact of the hyper-parameters $M, I$ and $S$, which control the behavior of PCGS.

\noindent\textbf{Impact of $M$: } 
$M$ is the threshold of the number of visits that we decide whether the topic distribution $\phiv_t$ of a topic is rapid-changing or slow-changing.
To investigate its effect, we run PCGS on the NYTimes dataset, setting $\betav=(1.0, 0.5, 0.25, 0.125)$,
and varying $M\in \{2^0, \dots, 2^{16}, \infty\}$, while tuning $\gamma$ to keep the number of topics around 500. PCGS becomes CGS when $M=\infty$, and approaches BGS when $M\rightarrow 0$. 
As shown in Fig.~\ref{fig:m}, the perplexity goes down and the time consumption goes up as $M$ grows.
We also find that $M\in [2^6, 2^9]$ provides a good trade-off between efficiency and quality. When $M=64$, there are 427 slow-changing topics which covers 99.7\% documents, so the change of rapid-changing topic counts (amount of communication) is kept small.

\noindent\textbf{Impact of $I$: } 
$I$ is the number of initializing iterations to sample from $p(\cv | \wv)$.
We run PCGS on the NYTimes (subset) dataset, setting $\beta_0=1$, and varying $I$ and $\gamma$. 
 It can be seen from Fig.~\ref{fig:iters} that the perplexity steadily decreases for large $I$, which again shows that our initialization strategy is helpful. We select a moderate $I=32$ for all the experiments. 

\noindent\textbf{Impact of $S$: } The hyper-parameter $S$ is the number of Monte-Carlo samples to approximate $p(\cv|\wv)$. When $S\rightarrow \infty$, we directly sample from $p(\cv|\wv)$ in the first $I$ iterations. We run PCGS on the NYTimes (subset) dataset, with $\betav=(1.0, 0.5, 0.25, 0.25)$ and $\gamma=10^{-40}$, and vary $S$ from 1 to 128. 
As shown in Fig.~\ref{fig:num-samples}, $S$ has little impact on both the number of topics and the perplexity, implying that a small $S$, e.g., $S=5$, is adequate.

\begin{figure*}[t]
\centering
\subfigure[NYTimes corpus]{
\includegraphics[width=0.16\linewidth]{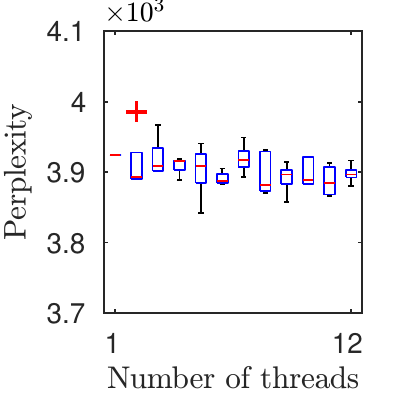}
\includegraphics[width=0.16\linewidth]{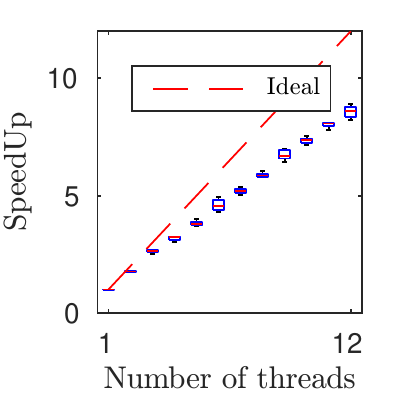}}
\subfigure[PubMed corpus]{
\includegraphics[width=0.16\linewidth]{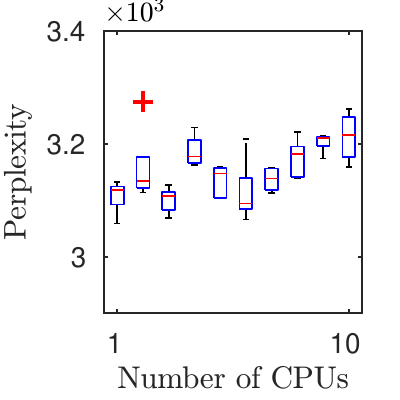}
\includegraphics[width=0.16\linewidth]{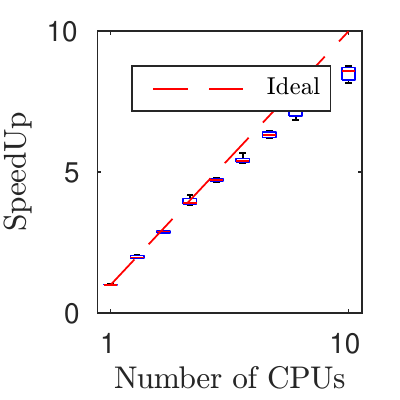}}
\subfigure[ClueWeb12 (small) corpus]{
\includegraphics[width=0.16\linewidth]{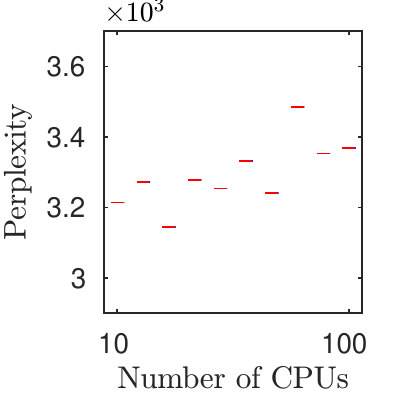}
\includegraphics[width=0.16\linewidth]{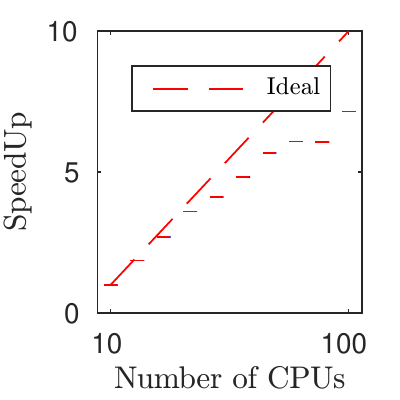}}\vspace{-.35cm}
\caption{Perplexity and speedup as the amount of computational resource increases. \label{fig:scalability}}\vspace{-.3cm}
\end{figure*}

\begin{figure*}[t]
\centering
\vspace{-.3cm}
\subfigure[Top 3 levels. The highlighted nodes are expanded as below.]{
\includegraphics[width=\linewidth]{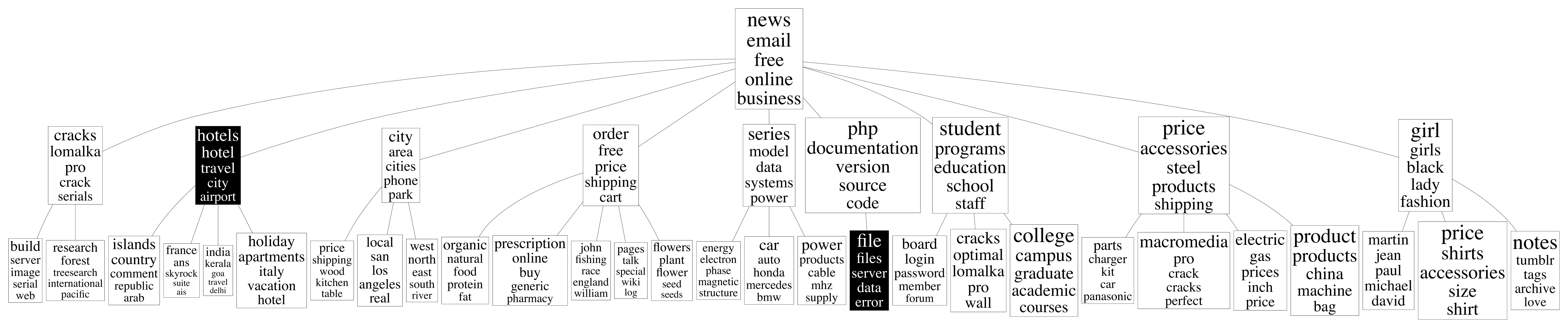}}\vspace{-.2cm}
\subfigure[Travel subtree]{
\includegraphics[width=0.49\linewidth]{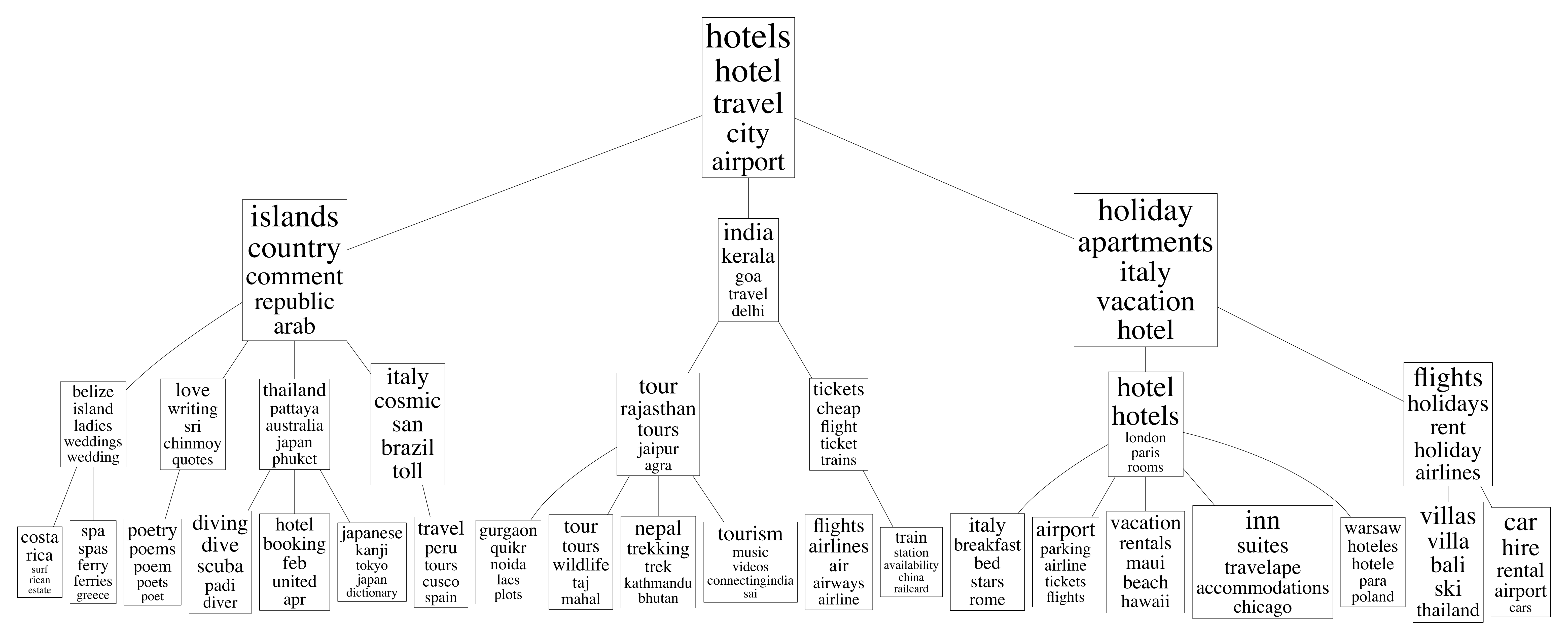}}
\subfigure[Computer subtree]{
\includegraphics[width=0.49\linewidth]{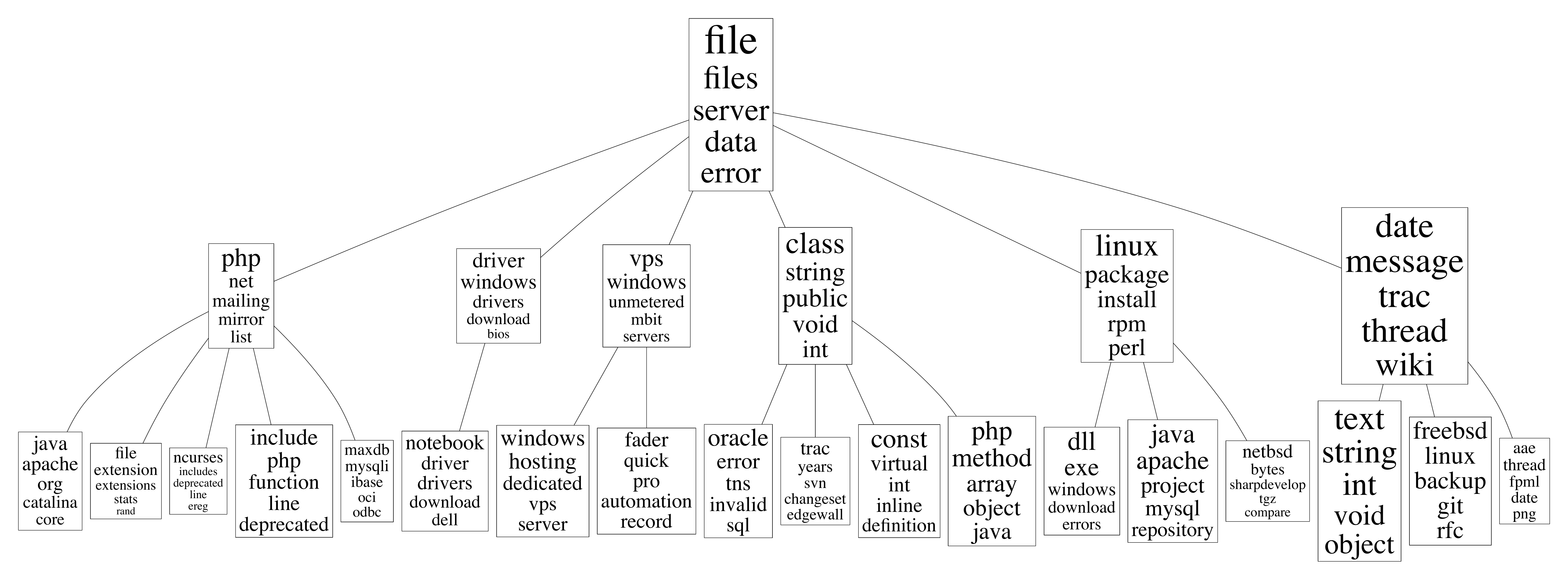}}\vspace{-.3cm}
\caption{Selected subtrees on the topic hierarchy extracted from ClueWeb12 (large). \label{fig:vis}}\vspace{-.2cm}
\end{figure*}

\subsection{Scalability}
Our experiments on scalability are in two folds: whether the quality of inference is affected by parallelization; and how good is the speedup. We first study the multi-thread setting, where the number of threads varies from 1 to 12 on the NYTimes corpus. The result is shown in Fig.~\ref{fig:scalability}(a),  where we observe that there is no apparent increase of perplexity as the number of threads grows. 
The speedup with 12 threads is 8.56. The probable reasons of imperfect speedup include serial region, contention for atomic variables, and limited memory bandwidth.

For the multi-machine setting, there are two CPUs per machine. We run our implementation on the PubMed corpus on 1 to 10 CPUs as shown in Fig.~\ref{fig:scalability}(b), and on the larger ClueWeb12 (small) corpus for 10 to 100 CPUs as shown in Fig.~\ref{fig:scalability}(c). The speedup is 8.5 from 1 to 10 CPUs, and 7.15 from 10 to 100 CPUs. The perplexity is slightly affected by parallelization when the number of CPUs exceeds 7 and 80 on the two datasets, respectively, indicating that the dataset is not large enough to utilize that many CPUs. 

Finally, to demonstrate the scalability, we learn a model with 1,722 topics of the 131-million-document ClueWeb12 (large) corpus with 50 machines, and the inference finishes in 7 hours. The results will be presented for qualitative evaluation in the next section.


\subsection{Qualitative Analysis}
\hyphenation{Clue-Web}
We now demonstrate the topic hierarchy obtained from the ClueWeb12 (large) corpus, which is a crawl of web pages. The corpus is obtained by tokenizing the original ClueWeb12 dataset, randomly selecting about 30\% documents, truncating the vocabulary size to 100,000 and keeping only the documents whose length is between $[50, 500]$. We show the selected parts of the obtained tree in Fig.~\ref{fig:vis}, where some topics whose number of occurrences does not pass a particular threshold are filtered out, and the font size of words is proportional to the 4th root of their frequency in the topic. The tree has 5 levels in total.\footnote{The visualization demo is available online at \url{http://ml.cs.tsinghua.edu.cn/~jianfei/scalable-hlda.html}.}

Fig.~\ref{fig:vis}(a) shows some selected topics on the first 3 levels. The root node contains the most commonly used words shared by all the documents. The second level contains a variety of general topics, such as ``software'', ``travel'' and ``city'', and the third level has more detailed concepts, e.g., the ``city'' topic on the second level splits as ``shopping'', ``city names'', and ``locations''. We further show the topic subtrees of all the layers rooted at the highlighted nodes to examine the fine-grained concepts. For example, in Fig.~\ref{fig:vis}(b) the ``travel'' topic is divided as ``islands'', ``India'' and ``vacations'', and the leaf level contains specific concepts, such as ``ferry'' and ``diving'', which are correctly placed under the ``islands'' topic. In Fig.~\ref{fig:vis}(c), the ``computer'' topic is divided as ``website'', ``windows'', ``vps'', ``programming'', ``linux'' and ``forum''. To our knowledge, this is the first time that hLDA is applied to large-scale web data, and the results demonstrate our ability on automatically learning topic hierarchy from web data.
\vspace{-0.1cm}
\section{Conclusions and Discussions}
We present a partially collapsed Gibbs sampling (PCGS) algorithm for the hierarchical latent Dirichlet allocation model, which is a combination of the collapsed weight algorithm and instantiated weight algorithm. 
The major feature of PCGS is that it is scalable and has high-quality inference. 
We also present an initialization strategy to further improve the model quality. To make PCGS scalable and efficient, we propose   vectorization and pre-processing techniques, concurrent data structures, and an efficient communication strategy.
The proposed algorithm and system are scalable to hundreds of millions of documents, thousands of topics, and thousands of CPU cores. 

In the future, we plan to extend our method to the more sophisticated nested HDP model~\cite{paisley2015nested,ahmed2013nested}. Developing sampling algorithms with sub-linear time complexity w.r.t. the number of topics via hashing is also an interesting direction.



\vspace{-0.1cm}
\appendix
\section{Derivation Details}
\subsection{Derivation of PCGS updates}
Rewrite the joint distribution in Sec.~\ref{sec:cgs} as:
$p(\wv, \zv, \thetav, \cv, \phiv_\mathcal C, \phiv_\mathcal I)$
=$\prod_{t\in \mathcal C}$
$p(\phiv_t)$
$\prod_{t\in \mathcal I}$ $p(\phiv_t) $
$\prod_{d=1}^D$ $p(\cv_d | \cv_{<d})$ $p(\thetav_d)$ 
$\prod_{d=1}^D$ $\prod_{n=1}^{N_d}$ $p(z_{dn}|\thetav)$ $p(w_{dn} | z_{dn}, \phiv_\mathcal C, \phiv_\mathcal I).$

Integrating out $\phiv_{\mathcal C}$ and $\thetav$, we have the marginal distribution
$p(\wv, \zv, \cv, \phiv_\mathcal I)$=$ 
\prod_{d=1}^D \big[ p(\cv_d | \cv_{<d}) \frac{B(\Cv_d+\alphav)}{B(\alphav)} \big]$
$\prod_{t\in\mathcal C} \frac{B(\Cv_t+\beta_{l_t})}{B(\beta_{l_t})}$
$\prod_{t\in\mathcal I}$ 
$\big[ \Dir(\phiv_t; \beta_{l_t})$
$\prod_{v=1}^V (\phi_{tv})^{C_{tv}}
\big].$

Utilizing the identity $\frac{B(\alphav+\ev_k)}{B(\alphav)}=\frac{\alpha_k}{\sum_k \alpha_k}$, where $\ev_k$ is a coordinate vector, we can derive the Gibbs sampling updates:

\textbf{Sample $\zv$:} Keeping only the terms  relevant with $z_{dn}$, we have 
$p(z_{dn}=l | w_{dn}=v, \wv, \zv_{\neg dn}, \cv, \phiv^{\mathcal I})$
$\propto$ $\frac{B(\Cv_d+\alphav)}{B(\alphav)}$
$\prod_{t\in\mathcal C}$ $\frac{B(\Cv_t+\beta_{l_t})}{B(\beta_{l_t})}$
$\prod_{t\in\mathcal I}$
$\prod_{v=1}^V \phi_{tv}^{C_{tv}}$
$\propto$ $\frac{B(\Cv_d^{\neg dn}+\Cv_d^{dn}+\alphav)}{B(\Cv_d^{\neg dn}+\alphav)}$
$\prod_{t\in\mathcal C}$ 
$\frac{B(\Cv_t^{\neg dn}+\Cv_t^{dn}+\beta_{l_t})}{B(\Cv_t^{\neg dn}+\beta_{l_t})}$
$\prod_{t\in\mathcal I}$
$\prod_{v=1}^V$ $\phi_{tv}^{C_{tv}^{\neg dn}+C_{tv}^{dn}}$
$\propto (C_{dl}^{\neg dn}+\alpha_l)$ 
$\begin{cases}
 \phi_{c_{dl}, v} & c_{dl} \in \mathcal I, \\
 \frac{C_{c_{dl},v}^{\neg dn}+\beta_{l_t}}{C_{c_{dl}}^{\neg dn}+V\beta_{l_t}} & c_{dl} \in \mathcal C.
\end{cases}
$


\textbf{Sample $\cv$:} 
$p(\cv_d=\cv | \wv, \zv, \cv_{\neg d}, \phiv^{\mathcal I})$
$\propto$
$p(\cv_d | \cv_{\neg d})$
$\prod_{t\in\mathcal C}$ $\frac{B(\Cv_t+\beta_{l_t})}{B(\beta_{l_t})}$
$\prod_{t\in\mathcal I}$ $\prod_{v=1}^V \phi_{tv}^{C_{tv}}$
$\propto$
$p(\cv_d | \cv_{\neg d})$
$\prod_{t\in\mathcal C}$ 
$\frac{B(\Cv_t^{\neg d}+\Cv_t^d+\beta_{l_t})}{B(\Cv_t^{\neg d}+\beta_{l_t})}$
$\prod_{t\in\mathcal I}$ $\prod_{v=1}^V$ $\phi_{tv}^{C_{tv}^d}$
$\propto$ 
$\mbox{nCRP}(\cv; \gammav, \cv_{\neg d})$ 
$\prod_{l=1}^L$
$\begin{cases}
f_I(d, c_{l}) & c_{l}\in \mathcal I,\\
f_C(d, c_l) & c_{l}\in \mathcal C.
\end{cases}
$


\textbf{Sample $\phiv^\mathcal I$:} For $t\in \mathcal I$, draw $\phiv_t \sim \Dir(\beta_{l_t} + \Cv_t)$.

\subsection{Derivation of computing $\log f_C(d, t)$}
We have 
$\log f_C(d, t)$ = $\log \frac{B(\Cv_t^{\neg d} + \Cv_t^{d}+\beta_{l_t})}{B(\Cv_t^{\neg d}+\beta_{l_t})}$
=$\big[ \sum_{v=1}^V \sum_{i=0}^{C_{tv}^d-1} \log(C_{tv}^{\neg d}+i+\beta_{l_t}) \big] + h_t $
=$\sum_{(v, o)\in \mathcal W^\prime_{dl}} 
\log (C_{tv}^{\neg d} + o + \beta_{l_t}) + h_t
$.

\begin{acks}
The work was supported by the National Basic Research Program (973 Program) of China (No. 2013CB329403), National NSF of China (Nos. 61620106010, 61322308, 61332007), the Youth Top-notch Talent Support Program, Tsinghua Tiangong Institute for Intelligent Computing, and Special Program for Applied Research on Super Computation of the NSFC-Guangdong Joint Fund (the second phase). J Lu and S Liu are supported by National NSF of China (No. 61672308).
\end{acks}

\bibliographystyle{ACM-Reference-Format}
\bibliography{scalable-hlda} 


\end{document}